\documentclass[11pt,twoside,twocolumn,a4paper]{article}

\usepackage{cvww}
\usepackage{times}
\usepackage{epsfig}
\usepackage{graphicx}
\usepackage{amsmath}
\usepackage{amssymb}
\usepackage{booktabs}
\usepackage{enumitem}
\usepackage{float}
\usepackage{capt-of}
\usepackage{tabu}
\usepackage[ampersand]{easylist}
\usepackage{subcaption}
\usepackage[export]{adjustbox}
\usepackage{siunitx}
\usepackage[bottom]{footmisc}

\usepackage[pagebackref=true,breaklinks=true,bookmarks=false]{hyperref}
\makeatletter
\DeclareRobustCommand\onedot{\futurelet\@let@token\@onedot}
\def\@onedot{\ifx\@let@token.\else.\null\fi\xspace}

\def\etal{\emph{et al}\onedot}

\cvwwfinalcopy 

\def\cvwwPaperID{24} 

\ifcvwwfinal\fi
\begin{document}

\title{Leveraging Outdoor Webcams for Local Descriptor Learning}

\author{Milan Pultar, Dmytro Mishkin, Ji\v{r}\'{i} Matas\\
Visual Recognition Group, Dept. of Cybernetics\\
Faculty of Electrical Engineering, CTU in Prague\\
{\tt\small milan.pultar@gmail.com, \{mishkdmy, matas\}@cmp.felk.cvut.cz}
}

\maketitle
\ifcvwwfinal\thispagestyle{fancy}\fi

\begin{abstract} 
We present AMOS Patches, a large set of image cut-outs, 
intended primarily for the robustification of trainable local feature descriptors to illumination and appearance changes. 
Images contributing to AMOS Patches originate from the AMOS dataset of recordings from a large set of outdoor webcams.

The semiautomatic method used to generate AMOS Patches is described. 
It includes camera selection, viewpoint clustering and patch selection.
For training, we provide both the registered full source images as well as the patches.

A new descriptor, trained on the AMOS Patches and 6Brown datasets, is introduced. 
It achieves state-of-the-art in matching under illumination changes on standard benchmarks.

\end{abstract}

\section{Introduction} 
\label{sec:introduction}

Local feature descriptors are widely used in tasks such as structure from motion \cite{schonberger2016structure, torii2018structure}, image retrieval \cite{shen2018matchable} and in applications like autonomous driving \cite{dewan2018learning}, which benefit from the robustness of the descriptors to acquisition conditions.

Recent years have witnessed a noticeable effort to move from handcrafted descriptors \cite{lowe2004distinctive} to those obtained by deep learning~\cite{mishchuk2017working,l2net}. Existing work explores possible architectures~\cite{tfeat2016, l2net}, loss functions~\cite{mishchuk2017working,He2018CVPR,Keller2018CVPR} and improvements of robustness to viewpoint changes by introducing large scale datasets from 3D reconstruction~\cite{mitra2018large, geodesc2018}.

Robustness to illumination and appearance changes has received little attention, yet it is a bigger challenge for modern descriptors~\cite{wxbs2015,balntas2017hpatches}. We tackle this problem by leveraging information from 24/7 webcams located worldwide~\cite{jacobs2007consistent,jacobs2009global}. 

We make the following contributions. First, we present a method for extracting veridical patch correspondences from the "static" cameras. Second, we present the AMOS Patches dataset\footnote{The dataset and contributing images are available at \url{https://github.com/pultarmi/AMOS_patches}}
for training of local feature descriptors with improved robustness to changes in illumination and appearance. 

\begin{figure}[]
    \centering
    \begin{subfigure}[c]{\linewidth}
      \frame{\includegraphics[width=\linewidth]{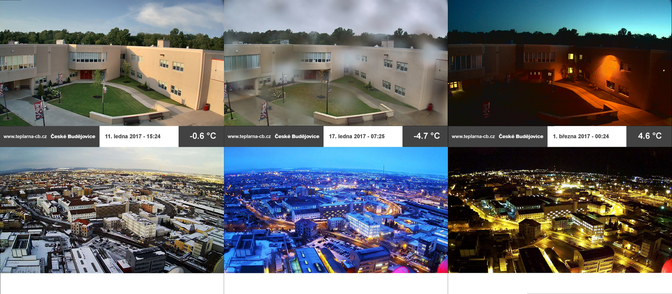}}
      \caption{}
    \end{subfigure}\hfill\par
    \begin{subfigure}[c]{\linewidth}
      \frame{\includegraphics[width=\linewidth]{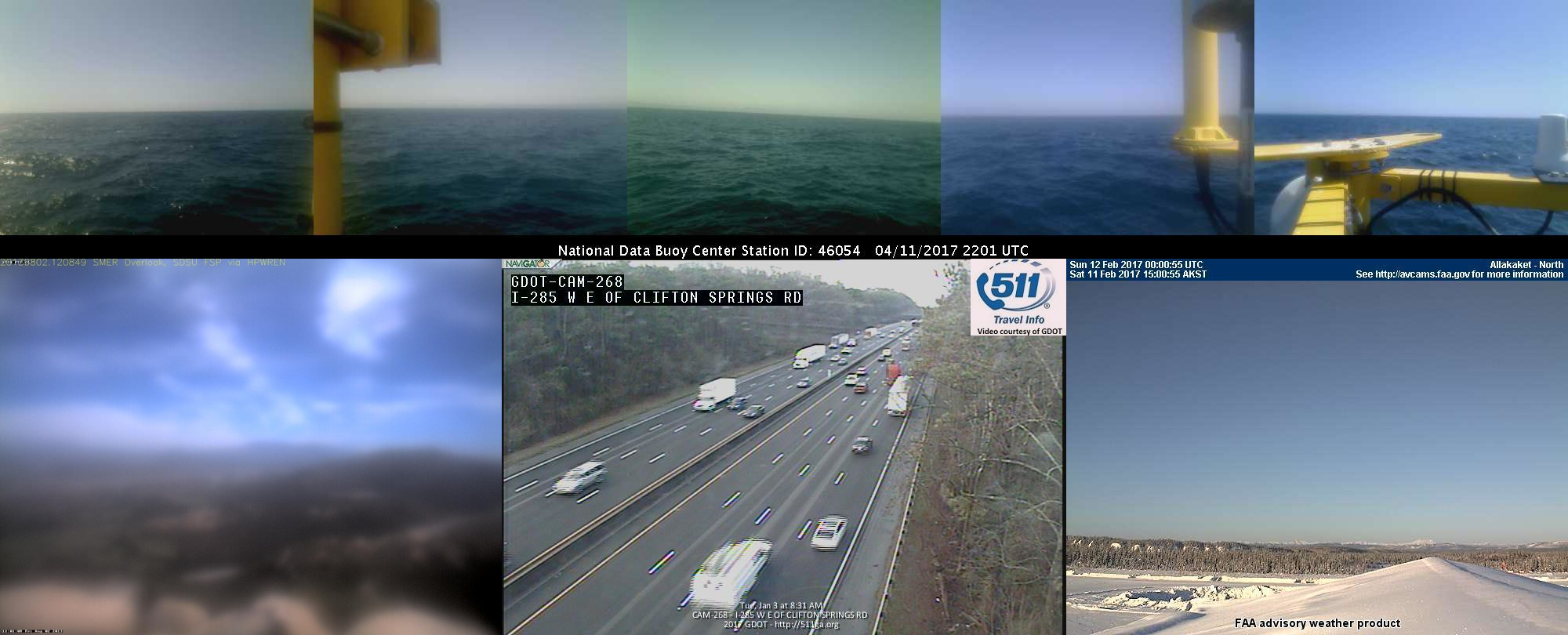}}
      \caption{}
    \end{subfigure}\hfill\par
    
    \caption{The AMOS dataset \cite{jacobs2009global, jacobs2007consistent} - example images from (a) cameras contributing to the AMOS patches set and (b) cameras unsuitable for descriptor training because of blur, dynamic content or dominant sky. }
    \label{fig:examples}
\end{figure}

As a final contribution, HardNet~\cite{mishchuk2017working} trained with AMOS Patches achieves state-of-the-art results in the commonly used HPatches benchmark~\cite{balntas2017hpatches}.

\section{Related Work} 
\label{sec:related_work}
The literature on local feature descriptors is vast. Here we focus on descriptors which are robust to illumination and appearance changes, refering the reader to Csurka~\etal~\cite{Csurka2018} for detailed survey on recent advances in local features. 
There are two main ways towards achieving robustness to illumination change: by descriptor construction and by learning on the appropriate dataset. 
Normalization of the patch mean and variance is a simple but powerful method, which is implemented in both SIFT~\cite{lowe2004distinctive} and modern learned descriptors~\cite{l2net, mishchuk2017working}.
The normalization makes the descriptor invariant to affine changes in pixel intensities in the patch. HalfSIFT~\cite{HalfSIFT2007} treats opposite intensity gradient directions as equal, trading off half of the SIFT dimensionality for being contrast reversal invariant. It works well in medical imaging and infrared-vs-visible matching.

The family of order-based descriptors like LIOP~\cite{LIOP2016} or MROGH~\cite{MROGH2011} operates on the relative order of pixel intensities in the patch instead of on the intensities themselves. Relative order (sorting) is invariant to any monotonically increasing intensity transformation. Descriptors like SymFeat~\cite{Hauagge2012}, SSIM~\cite{Shechtman2007} and learned DASC~\cite{DASC2017} encode local symmetries and self-similarities. 
Another possibility is, instead of constructing a descriptor, to apply some transformation to the pixel intensities as done by the learned RGB2NIR~\cite{Zhi_2018_CVPR} or hand-crafted LAT~\cite{LAT2017}, and then use a standard descriptor, e.g.\ SIFT. 

Data-driven approaches mostly include Siamese convolution networks with modality-specific branches, like the Quadruplet Network~\cite{Qnet2017}. The decision which branch to use for a specific patch comes from an external source or a domain classifier. HNet~\cite{HNet2018} uses an auto-encoder network and style transfer methods like CycleGAN~\cite{CycleGAN2017} for emulating different modalities. 

There is a number of image-level datasets specifically designed for testing illumination-robust recognition: DTU Robot~\cite{dturobot2012}, OxfordAffine~\cite{Mikolajczyk2005}, RobotCar dataset~\cite{RobotCarDatasetIJRR}, Aachen Day-Night~\cite{Sattler2018}, GDB~\cite{yang2007registration}, SymBench~\cite{Hauagge2012}, etc. Despite the importance of the topic, the number of patch-level datasets for illumination-robust descriptors is small, especially those which are suitable for descriptor learning.  To our best knowledge, Two Yosemite sequences from the Phototour dataset~\cite{brown2007automatic} and the Illumination split of the HPatches dataset~\cite{balntas2017hpatches} are the only ones suitable for descriptor learning and are publicly available.

\section{Creating AMOS Patches} 
\label{sec:creating_patches}

\begin{figure*}
    \centering
    \includegraphics[scale=0.16, width=\textwidth]{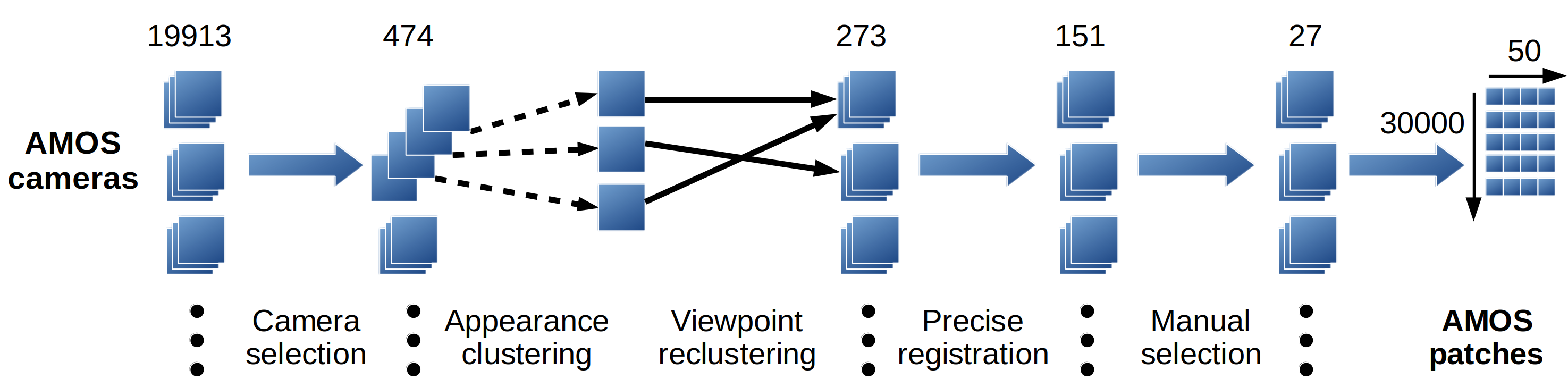}
    \caption{The pipeline of AMOS patches consists of: camera selection to filter out dynamic or empty scenes, appearance clustering to remove redundant images, viewpoint reclustering to tackle switching cameras, precise registration for further filtering, manual pruning for final selection of views and patch sampling.}
    \label{fig:pipeline}
\end{figure*}

AMOS \cite{jacobs2009global, jacobs2007consistent} is a continuously growing publicly available dataset collected from outdoor webcams, currently containing over one billion (or 20 TB) images. It is organized into individual camera directories, which are split into folders according to the year and month of the acquisition. The size of the images varies, and so does their quality and the number of images in each camera directory. A typical AMOS camera is static and has approximately 300 times 300 pixel size. Many cameras store images in all seasons and during the whole day.

The advantage of static cameras lies in the fact that they show the same structure under different weather and lighting conditions. Therefore, if observing a static scene, they are highly suitable for training of local feature descriptor robust to illumination and appearance changes.

We learned the hard way that using this type of data is not trivial. Firstly, due to the dataset size, it is not feasible with moderate computing power to load such data into memory. Moreover, preprocessing would take a prohibitive amount of time. Secondly, the training procedure is sensitive to misregistration of the images and the presence and size of moving objects. Many cameras experience technical issues such as: being out of focus, rotating over time, displaying highly dynamic scene (e.g. sky, sea waves), which all significantly hurt the performance of the trained descriptor, as discussed later.
 
Therefore, we developed a pipeline for the creation of AMOS Patches, shown in Figure~\ref{fig:pipeline}, which entails several steps to create a clean dataset with veridical patch correspondences. These methods focus on the selection of cameras and images, detection of view switching in a camera and the registration of images. Because of several not easily detectable problems, it was still necessary to perform final manual check of the selected image sets.

\subsection{Camera selection} 
The first step --- camera selection --- aims at choosing a subset of cameras which are suitable for training, i.e.\ do not produce very dark images,  are sharp and do not display moving objects like cars or boats.

The procedure uses two neural networks, a sky detector \cite{mihail2016sky} and an object detector \cite{torchcv}, and computes simple statistics for each of 20 randomly chosen images in each camera directory.

The camera selection took approximately one week on a single PC (Intel\textsuperscript{\textregistered} Xeon\textsuperscript{\textregistered} CPU E5-2620) with one GPU GTX Titan X. Processing more images by the neural network detectors increases both the precision of the method and the running time. Our choice is therefore based on the available computation power.

Each image is then checked whether it satisfies the following conditions:

\begin{itemize}
\item $f_{1}~:$ sky area $<50\%$ \hfill \textit{not empty}
\item $f_{2}~:$ no detected cars or boats \hfill \textit{not dynamic}
\item $f_{3}~:$ $\mathrm{Var}(\nabla^{2}$ \text{image}) $\geq 180$ \hfill \textit{sharp}
\item $f_{4}:$ mean pixel intensity $>30$  \hfill \textit{not black}
\item $f_{5}:$ image size $ >(700,700)$  \hfill \textit{large}
\end{itemize}

A camera is kept if at least 14 out of the 20 images pass the check.

The filter $f_5$ is the most restrictive, it removes $91\%$ of the cameras -- AMOS contains mostly low resolution images. The reasoning behind using $f_5$ is that images with smaller size often observe a motorway or are blurred. Also, such cameras would not generate many patches. We want to select only a relatively small subset of the cameras with the predefined characteristics and therefore an incorrect removal of a camera is not a problem.

Several cameras were removed because of corrupted image files. The resulting set contains 474 camera folders which were subject to subsequent preprocessing.

\subsection{Appearance clustering by K-means} 
The resulting data is of sufficient quality, but it is highly redundant: images shot in 10 minute intervals are often indistinguishable and very common. To select sufficiently diverse image sets, we run the K-means clustering algorithm with $K$=120 to keep the most representative images. We use the fc6 layer of the ImageNet-pretrained AlexNet~\cite{krizhevsky2012imagenet} network as the global image descriptor. While not being the state-of-the-art, AlexNet is still the most effective architecture in terms of speed~\cite{Canziani2016}, with an acceptable quality.

At this stage of the pipeline, there are $K$=120 images for each of the $C$=474 cameras selected, a feasible number for training with the computational resources available.

Feature descriptor training with patches selected from this image set was not successful.
We were unable to achieve accuracy higher than 49.1 mean average precision (mAP) in the HPatches matching task; 
the state-of-the-art is 59.1 mAP -- GeoDesc~\cite{geodesc2018}.

\subsection{Viewpoint clustering with MODS} 
After examining the data closely, we found that many of the cameras switch between a few views, which breaks our assumption for the generation of ground truth correspondences via identity transformation. In order to filter out the non-matching views, we run MODS \cite{mishkin2015mods}, a fast method for two-view matching, and split each camera folder into clusters, called views, by applying a threshold on the number of inliers and the difference between the homography matrix and the identity transform.

Let $(x_1, x_2, ..., x_K)$ be a set of images in a camera folder in arbitrary order. MODS matching is first run on pairs $(x_1,x_2),(x_1,x_3),...(x_1,x_K)$. Image $x_1$ becomes the reference image in a newly created view, which contains $x_i$ for which the registration yields more than 50 inliers and SAD($H(x_1, x_i), I_3) < 50$. SAD denotes the sum of absolute differences, $H$ denotes a homography matrix normalized by the element in position $(3,3)$, $I_3$ is 3x3 identity matrix. All images in the created view are then removed from the processed image set. The step is repeated until no images remain.

We observed that the number of the resulting views in one camera folder depends on phenomena other than camera movement. For example, in cases where there is a fog or very rainy weather, MODS fails to match most of the image pairs and many of them form a single element cluster, which is excluded from further processing. For each camera, we keep only the view with the largest number of images, if it has more than 50. Each remaining view is reduced to 50 images by random selection.

\subsection{Registration with GDB-ICP} 
\begin{figure}[]
    \centering
    \includegraphics[max width=0.5\textwidth]{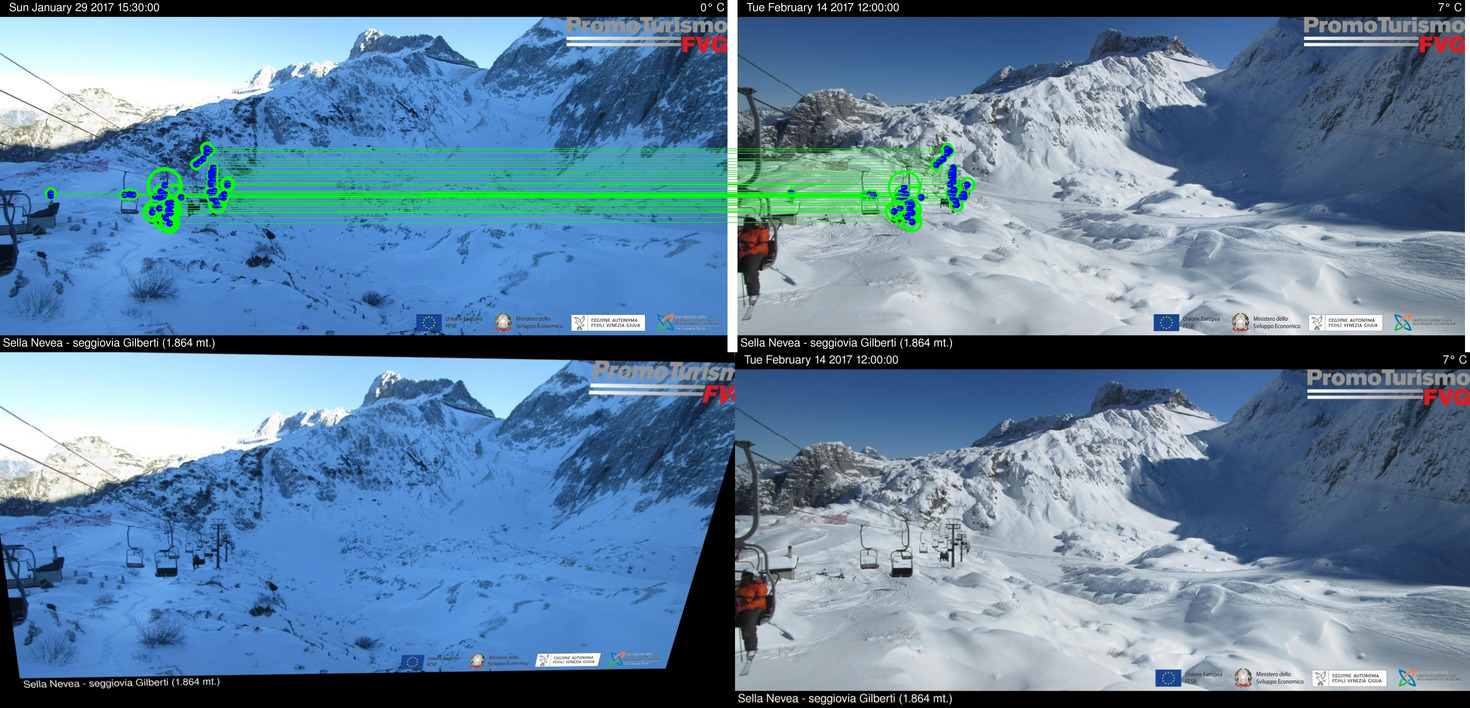}
    \caption{MODS registration failure, most of the correspondences are on moving structures.
     Top: an image pair with marked inliers. Bottom: wrongly transformed image (left) and the reference.}
    \label{fig:mods_wrong}
\end{figure}

While the MODS method is effective in matching and subsequent reclustering of camera sequences, in most of the cases the estimate of the global homography is not suficiently precise. MODS often outputs a homography valid for only small area in the image, see the example shown in Figure~\ref{fig:mods_wrong}. Therefore, the views contain also images which are not correctly aligned. To alleviate the problem, we run Generalized Dual Bootstrap-ICP \cite{yang2007registration} to prune the set of views, keeping those where this second registration is successful.

The registration proceeds as follows. Each view folder contains images $(x_1, x_2, ..., x_{50})$, where image $x_1$ is the MODS reference. The GDB-ICP registration is run on pairs $(x_1, x_2), (x_1, x_3),... (x_1, x_{50})$ and warped images $x'_2, x'_3, ..., x'_{50}$ are obtained. If registration fails on any pair, the whole view is removed.

After the precise registration with GDB-ICP, 151 views remained. 
It is feasible to manually inspect such a set.

\subsection{Manual pruning} 
\begin{figure}[]
    \centering
    \includegraphics[max width=0.5\textwidth]{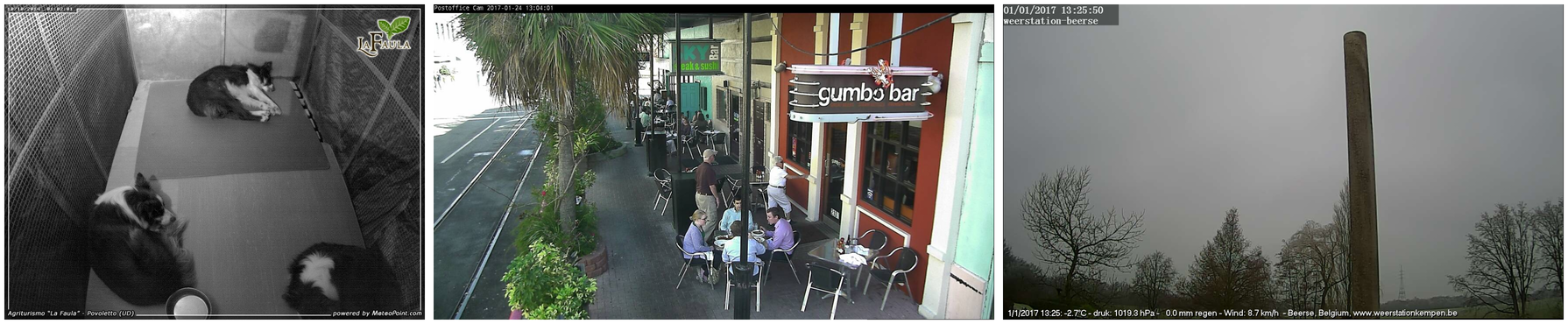}
    \caption{Manually pruned views. Examples of dynamic scenes (left, center) and a cloud-dominated scene not removed during camera selection (right). }
    \label{fig:manual_images}
\end{figure}

A few problems remain, see Figure \ref{fig:manual_images}, such as dynamic scenes,
undetected sky (the sky detector fires mostly on the clear blue sky). As a precaution, we also removed views with very similar content and views from different cameras observing the same place from a different viewpoint. 
We tried to use the scene segmentation network \cite{zhou2018semantic} to detect moving objects, but the result was not satisfactory. The final selection is therefore done by hand, resulting in a set of 27 folders with 50 images each.

\subsection{Patch selection, training} 
The last phase of the AMOS Patches pipeline consists of sampling images to obtain patch centers, scales and angles, and subsequent cropping from source images. We tested two approaches.
First, one may average the images in a view and evaluate a response function over the resulting image. Second, one may evaluate the response function over all images in a view and average the outputs. The resulting 2D map is then used as a probability mask for the selection of patch centers. Scales and angles are sampled independently at random from a predefined range.

For training, we use the hard-in-batch triplet margin loss~\cite{mishchuk2017working}. This structured loss requires corresponding (positive) pairs of patches on input. Therefore, AMOS Patches dataset consists of sets of patches cropped from the same position in each image in a view. The size of each patch set is equal to the number of images in a view directory, which is 50 in our case. Each patch is resampled to 96 times 96 pixels.

During training, we apply random affine transformation and cropping to get patches of smaller size. First, random rotation from range $(\ang{-25},\ang{25})$, scaling from range $(0.8, 1.4)$ and shear are applied. Second, from a 64 times 64 center of a patch we crop a 32 times 32 region with random scale. These transformed patches are the input for training.

We use the HardNet implementation in Pytorch \cite{paszke2017automatic}. For training we use batch size of 1024, 20 epochs, learning rate = 20, SGD optimizer with momentum $= 0.9$.

\section{Evaluating influences on precision} 
\label{sec:influences}
We examine the influence of several choices made before and during training. They relate to batch formation, patch selection and the dataset size. Also, we show the importance of registration of images in a view.

Two evaluation tasks are considered. In the matching task, there are two equally sized sets of patches from two different images. The descriptor is used to find a bijection between them. The average precision (AP) over discrete recall levels is evaluated for each such pair of images. Averaging the results over a number of image pairs gives mAP (mean AP). In the verification task there is a set of pairs of patches. The descriptor assigns a score that the two patches in a pair correspond. Precision-recall curve is then plotted based on the sorted (according to the score) list of patch pairs distances.

\subsection{Registration} 
\begin{figure}[]
    \centering
    \includegraphics[max width=0.5\textwidth]{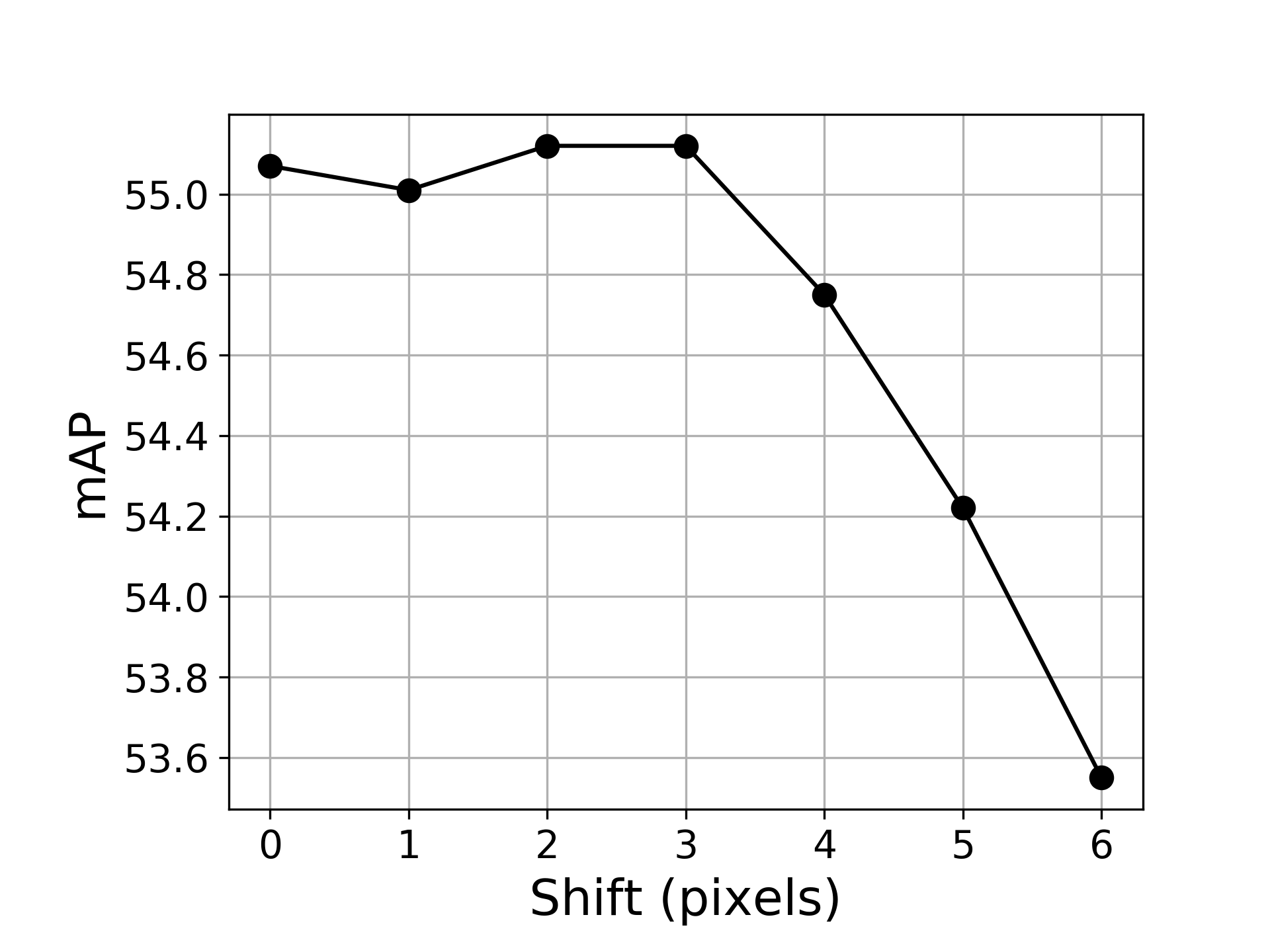}
    \caption{HPatches matching. The mAP score of Hardnet trained on AMOS patches displaced by different shifts.}
    \label{fig:deregistration}
\end{figure}

In this experiment we show the importance of the precise alignment of images. We displace each patch by different shifts and observe the influence on the HPatches matching score, see Figure \ref{fig:deregistration}. Notice how the performance of the descriptor improves with a small shift, but then quickly deteriorates. We use $\#$source views $=27$ (all), 30000 patch sets and Hessian weighting without averaging. These parameters are defined below.

\subsection{Number of source views} 
\begin{figure}[]
    \centering
    \includegraphics[max width=0.5\textwidth]{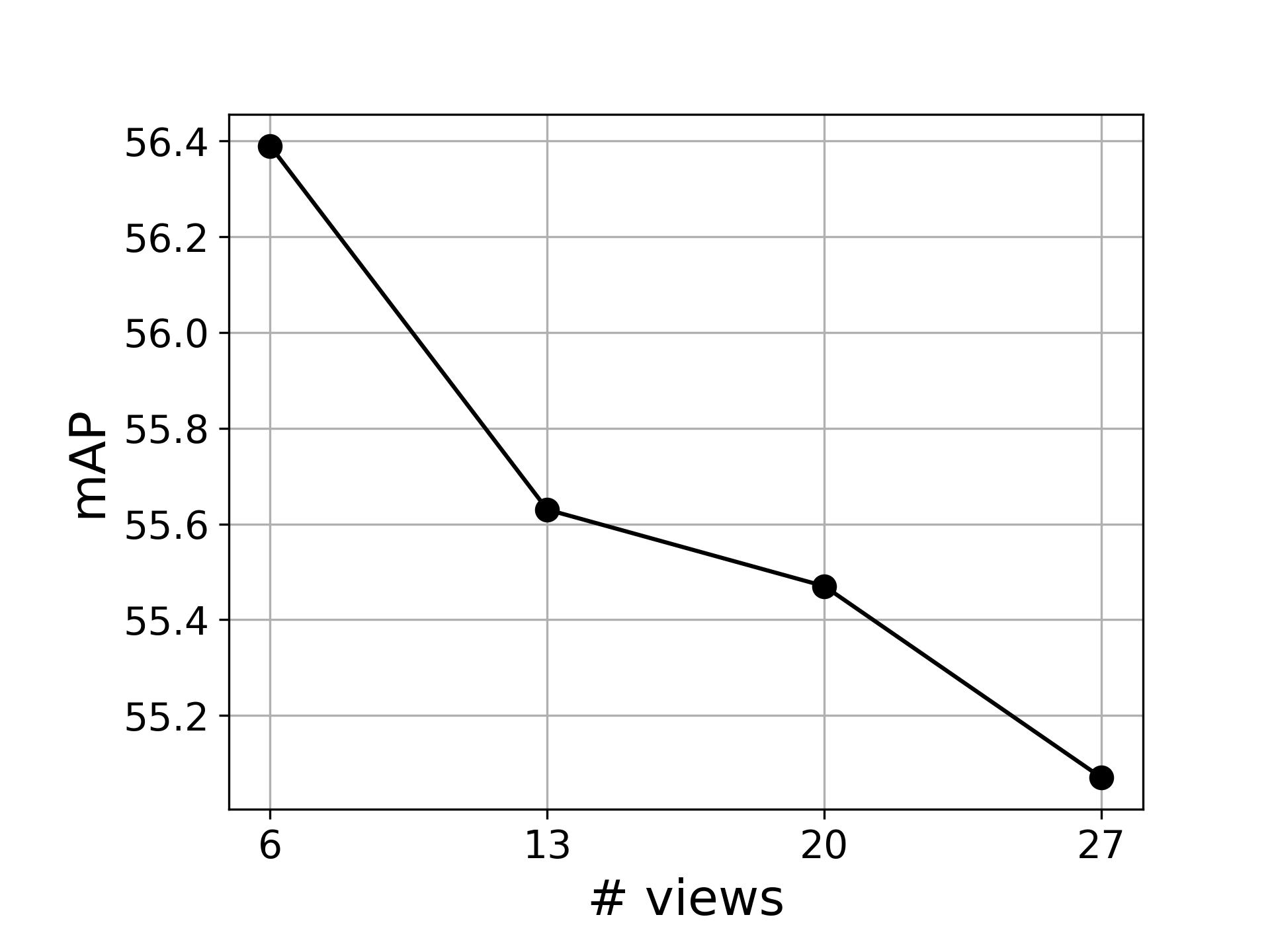}
    \caption{HardNet mAP score in HPatches matching task as a function of the number of source views for a batch. Views are selected randomly in each iteration. Dataset consists of 27 views in total.}
    \label{fig:n_cams}
\end{figure}

The hard-in-batch triplet margin loss is influenced by the composition of a batch. This experiment shows that lowering the number of views from which we choose patches to form a batch is an effective way to improve training on AMOS Patches, see Figure \ref{fig:n_cams}.
We interpret this behaviour as follows. Reducing the number of views increases the number of negative patches from the same scene, which are often the most difficult to distinguish. 

\subsection{AMOS Patches size} 
\begin{figure}[]
    \centering
    \includegraphics[max width=0.5\textwidth]{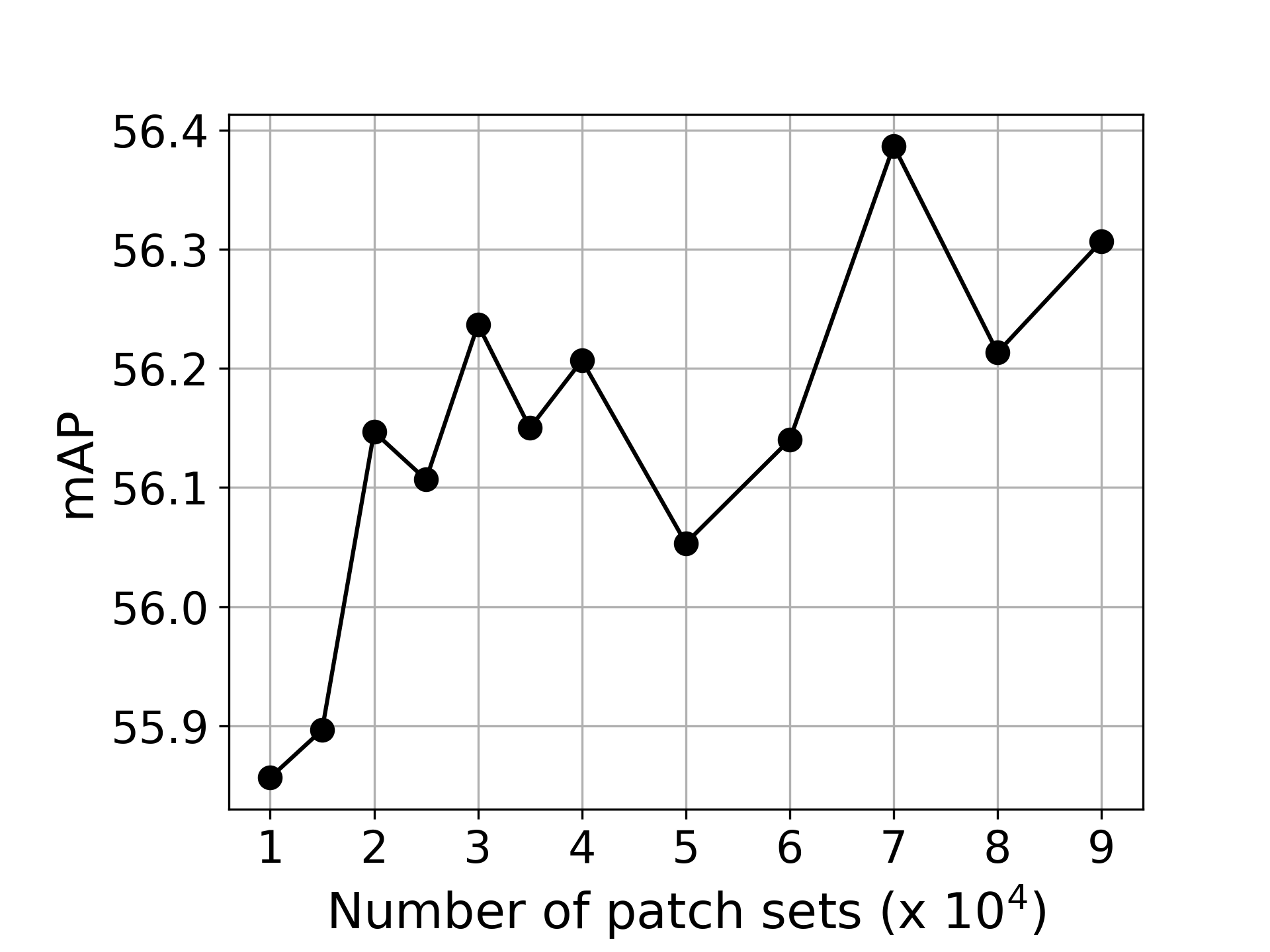}
    \caption{HardNet mAP score in HPatches matching task evaluated for different sizes of AMOS patches training dataset. Each value is an average over 3 different randomly generated datasets of the same size.}
    \label{fig:n_patches}
\end{figure}

Here we examine the influence of the dataset size, i.e.\ the number of patch sets created from source views, see Figure \ref{fig:n_patches}. We use the results from the previous experiment and choose \#(source cameras) $=$ 6. The graph shows there is a rough increase in HPatches matching score on bigger datasets.
Based on the result, we fix the number of patches to be 30 000 to trade off dataset compactness for slightly higher performance.

\subsection{Patch sampling} 
The patch selection method is partially determined by two independent choices: the response function and the averaging method. First, we find the best response function (Table \ref{tab:response}), then we keep it fixed and determine the optimal averaging function, which may apply either to outputs from the response function (Table \ref{tab:averaging}) or to images in a view (Table \ref{tab:source_img}).

\begin{table}[htb]
   \centering
   \caption{Patch sampling: Influence of the response function on HPatches matching score (mAP).}
   \begin{tabular}{lc}
\toprule
    Weighting  & mAP\\
\midrule
   Uniform  & 56.20 \\
   Hessian   & 56.39 \\
  $\textstyle\sqrt{\text{Hessian}}$& \textbf{56.49} \\
 NMS($ \textstyle\sqrt{\text{Hessian}}$)  & 56.18 \\
 \bottomrule
   \end{tabular}
   \label{tab:response}
\end{table}

\begin{table}[htb]
   \centering
   \caption{Patch sampling: Influence of the response averaging on HPatches matching score (mAP). Weighting function is $ \textstyle\sqrt{\text{Hessian}}$.}
   \begin{tabular}{lc}
\toprule
   Averaging & mAP\\
\midrule
none & \textbf{56.49} \\
mean  & 56.10 \\
median & 56.45 \\
 \bottomrule
   \end{tabular}
   \label{tab:averaging}
\end{table}

\begin{table}[htb]
   \centering
   \caption{Patch sampling: Influence of the image averaging on HPatches matching score (mAP). Weighting function is $ \textstyle\sqrt{\text{Hessian}}$.}
   \begin{tabular}{lc}
\toprule
   Image & mAP\\
\midrule
random & 56.49 \\
median  & 56.44 \\
mean & \textbf{56.58} \\
 \bottomrule
   \end{tabular}
   \label{tab:source_img}
\end{table}


\section{Evaluation} 
\begin{figure}[]
    \centering
    \includegraphics[max width=0.5\textwidth]{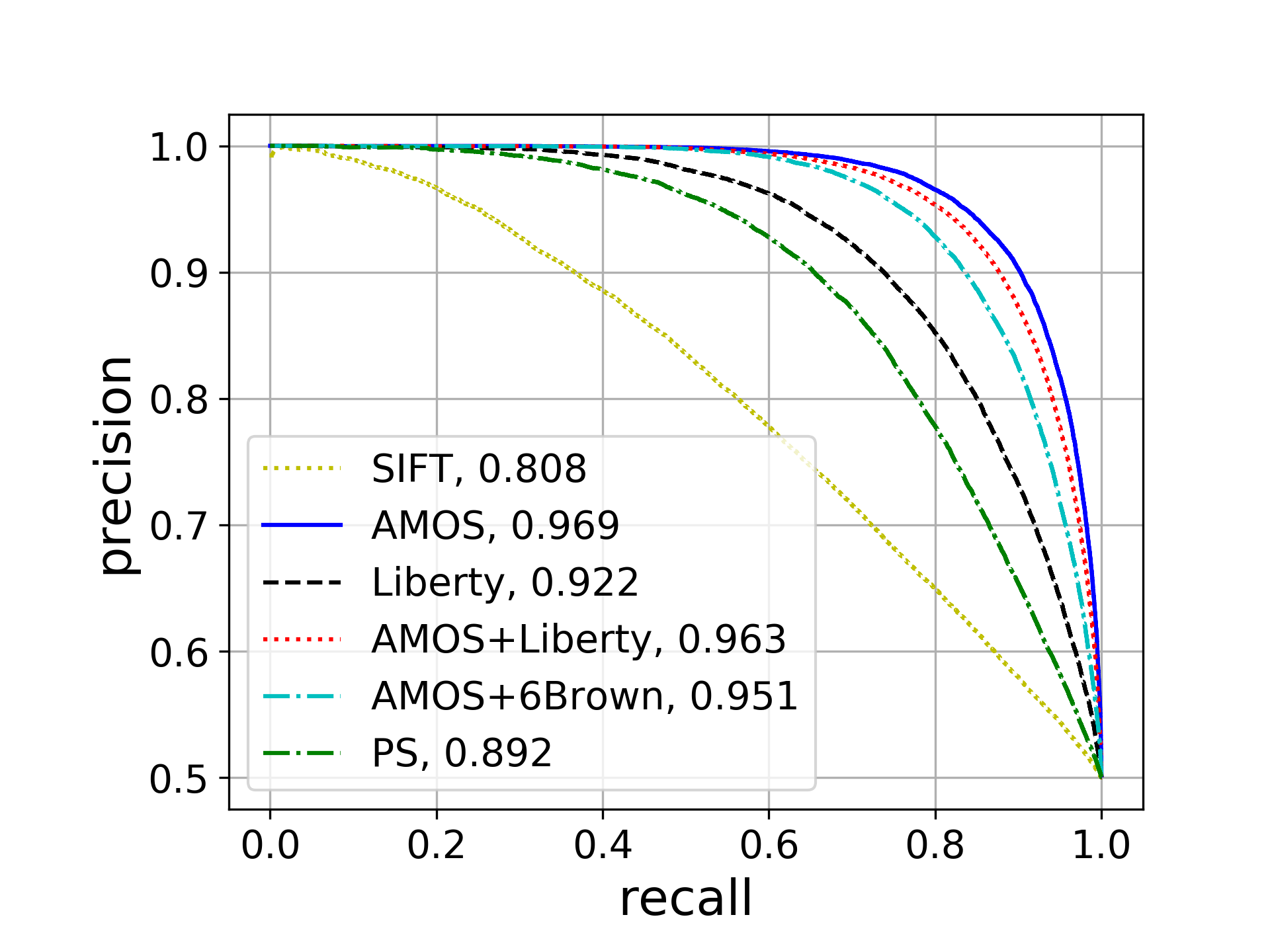}
    \caption{HardNet performance on the AMOS test set, when trained on the AMOS, Liberty, AMOS and Liberty, AMOS and 6Brown and PS~\cite{mitra2018large} datasets. SIFT results are provided as a baseline.}
    \label{fig:amos_test}
\end{figure}

\begin{figure*}
    \centering
    \begin{subfigure}[c]{0.5\linewidth}
      \includegraphics[width=\linewidth]{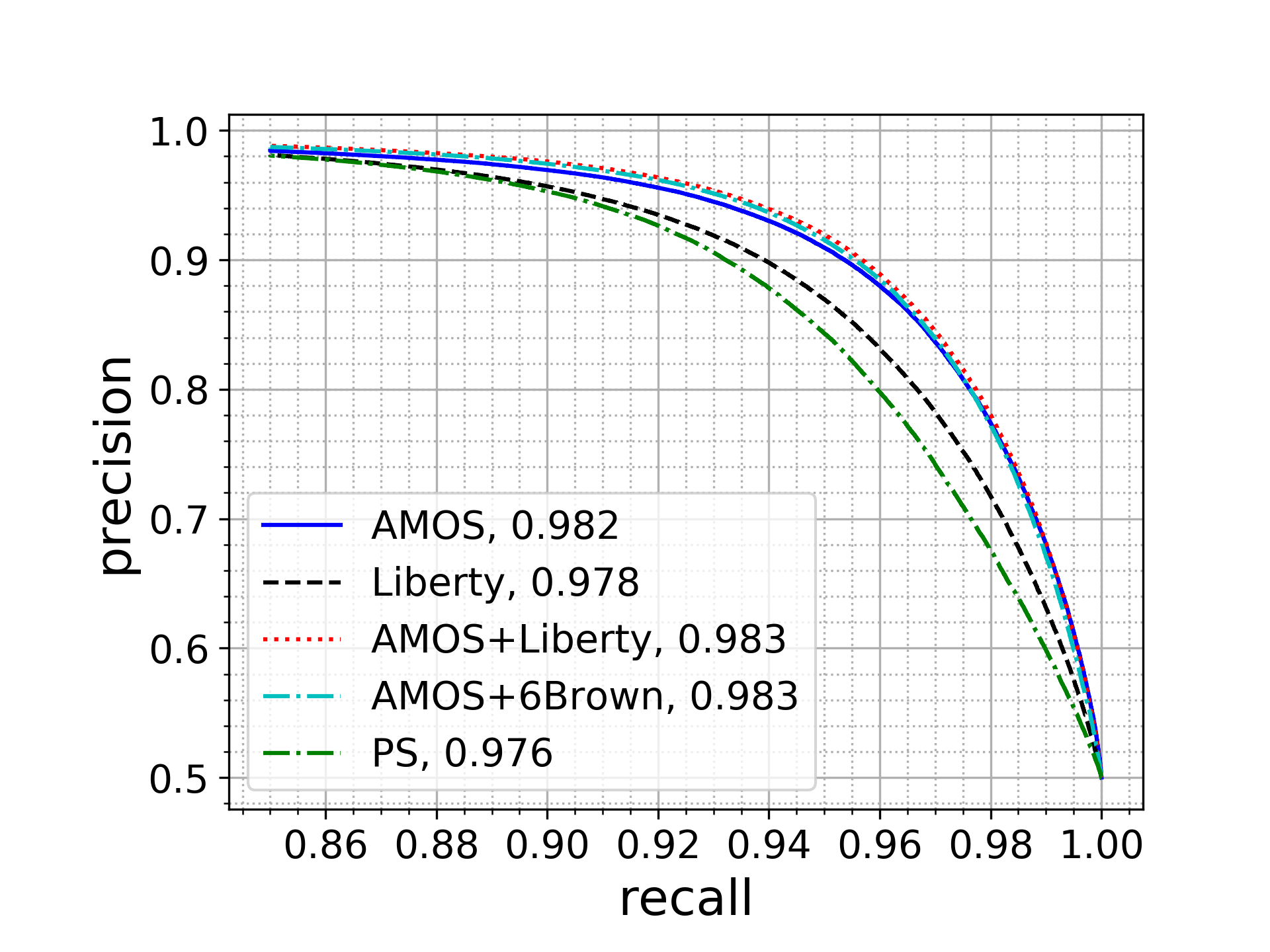}
      \caption{HPatches intra easy}
    \end{subfigure}\hfill
    \begin{subfigure}[c]{0.5\linewidth}
      \includegraphics[width=\linewidth]{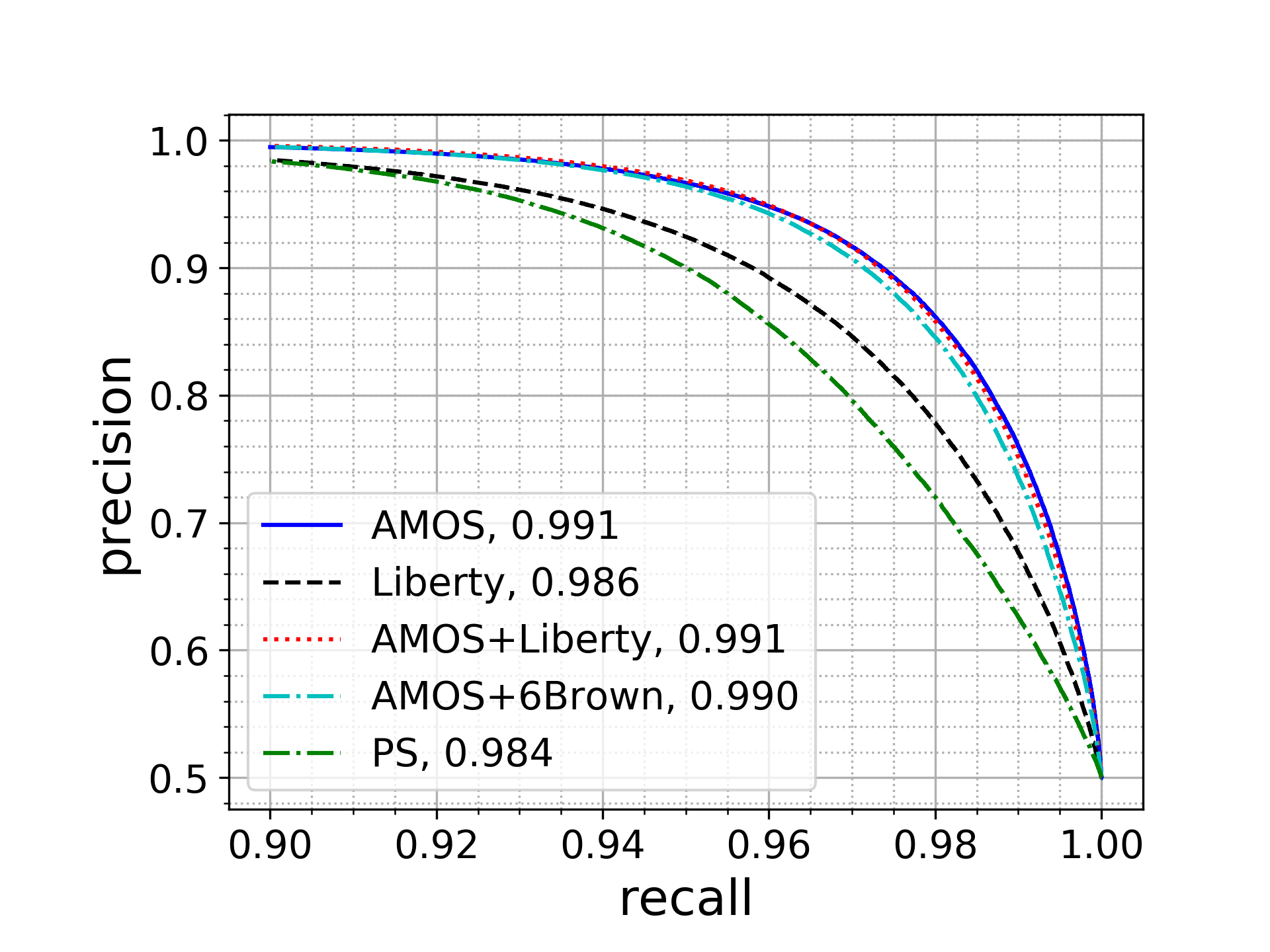}
      \caption{HPatches inter easy}
    \end{subfigure}\hfill\par
    \medskip
    \begin{subfigure}[c]{0.5\linewidth}
      \includegraphics[width=\linewidth]{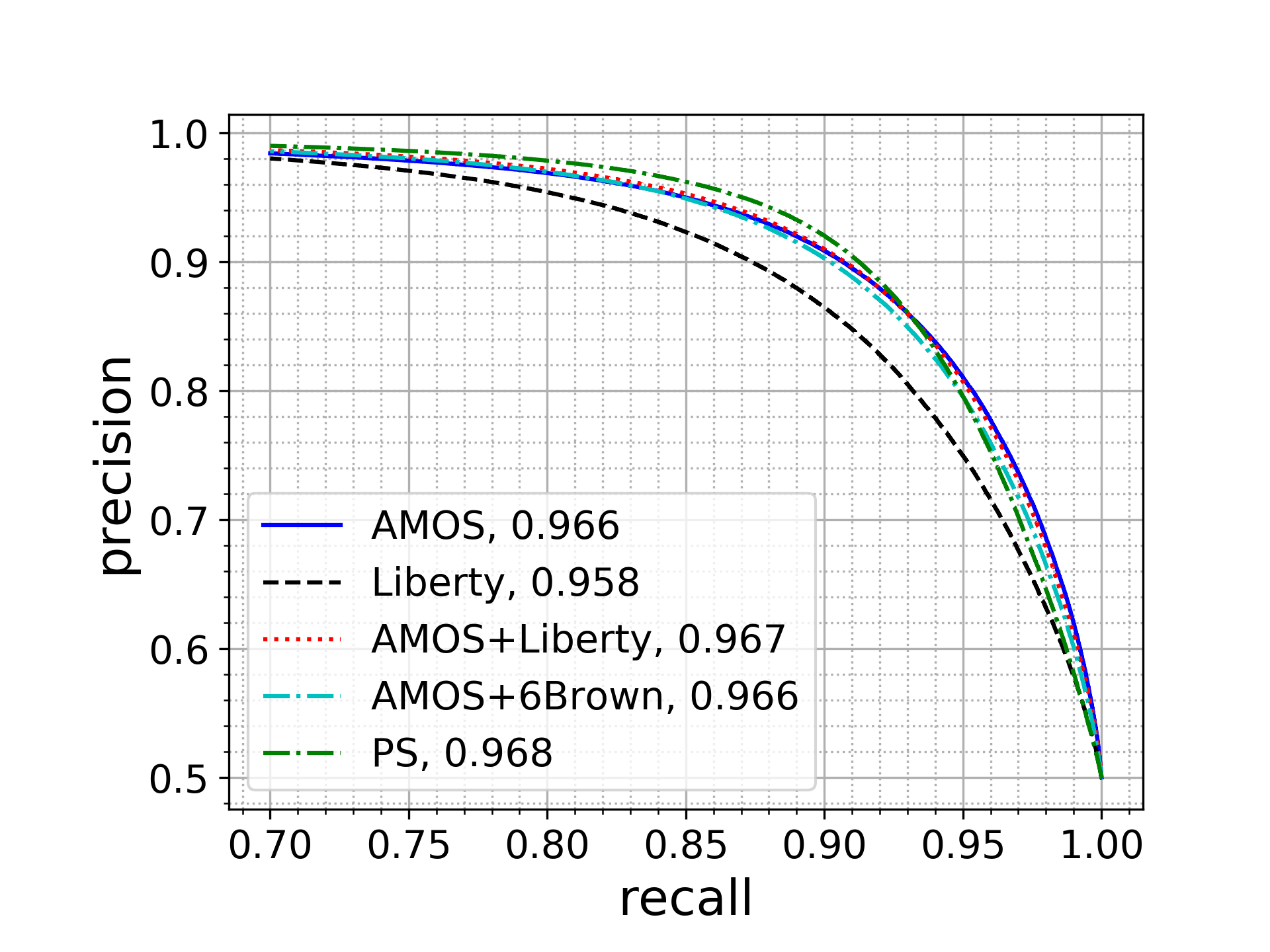}
      \caption{HPatches intra hard}
    \end{subfigure}\hfill
    \begin{subfigure}[c]{0.5\linewidth}
      \includegraphics[width=\linewidth]{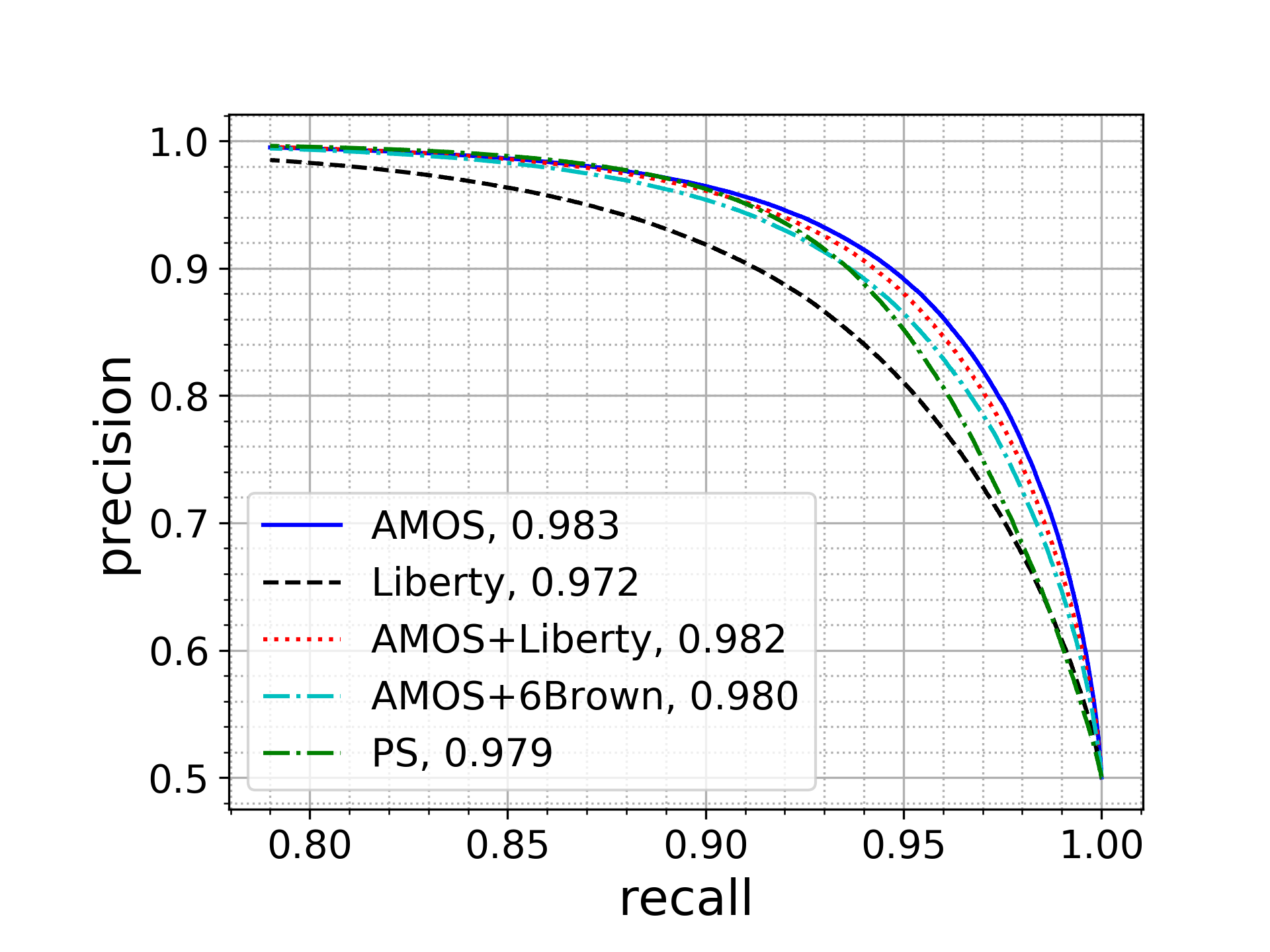}
      \caption{HPatches inter hard}
    \end{subfigure}\hfill\par
    \medskip
    \begin{subfigure}[c]{0.5\linewidth}
      \includegraphics[width=\linewidth]{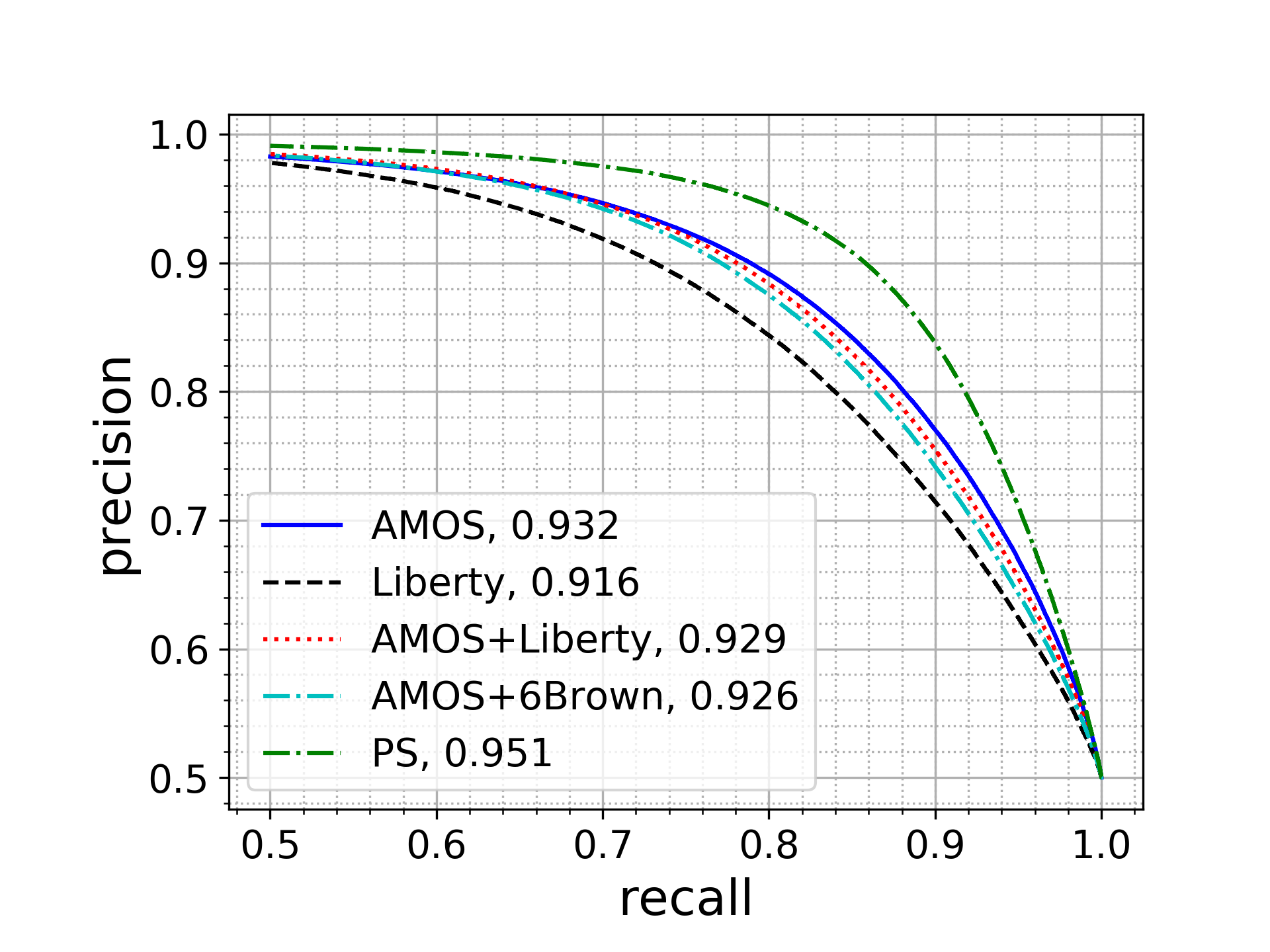}
      \caption{HPatches intra tough}
    \end{subfigure}\hfill
    \begin{subfigure}[c]{0.5\linewidth}
      \includegraphics[width=\linewidth]{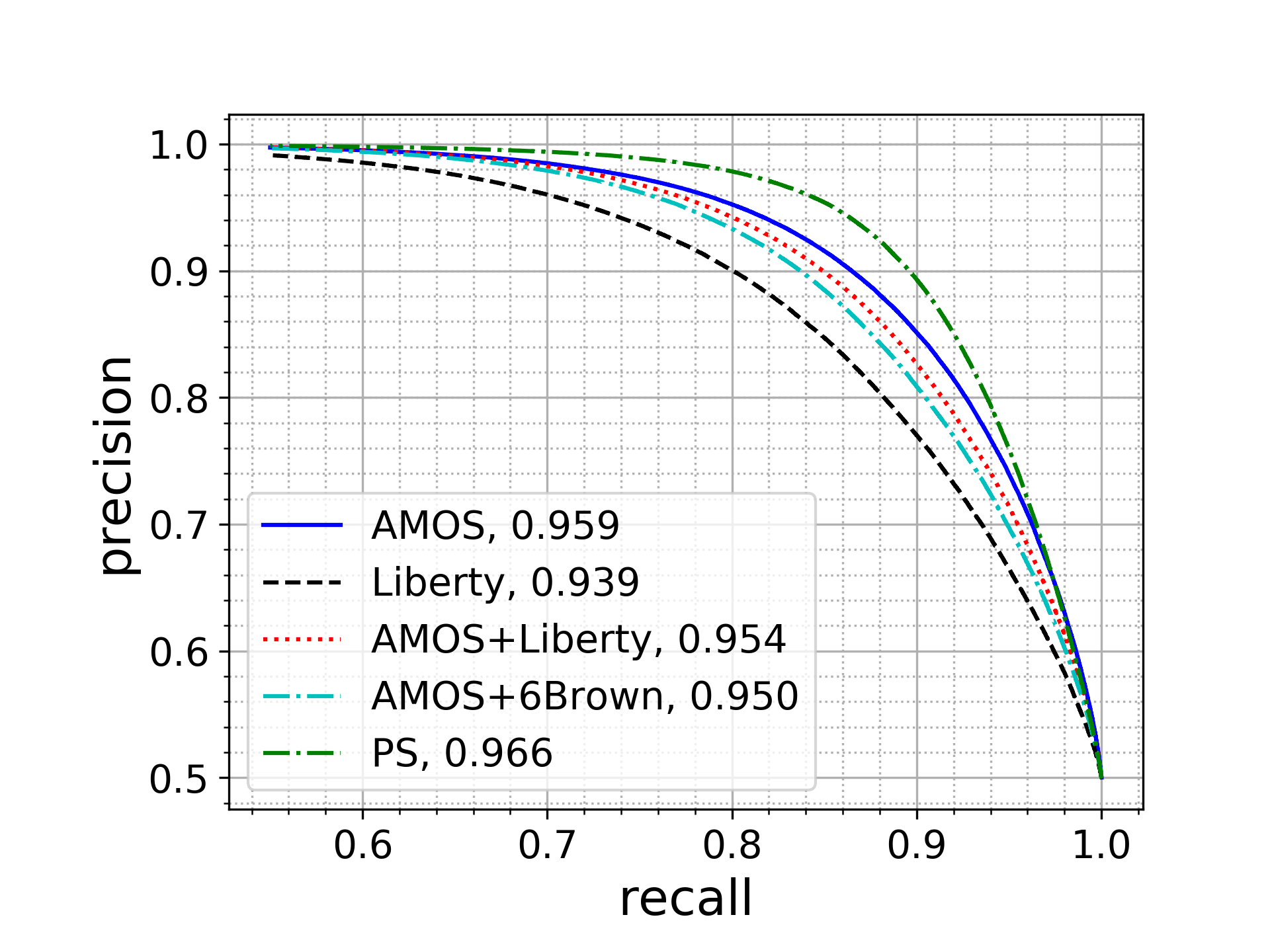}
      \caption{HPatches inter tough}
    \end{subfigure}\hfill\par
    \caption{HardNet performance evaluated on the HPatches benchmark. Precision-recall curve is presented based on the output from the verification task. Legend shows the training set name with the corresponding AUC. }
    \label{fig:verification}
\end{figure*}

\paragraph{HPatches and AMOS benchmarks.} The evaluation shows that HardNet trained on AMOS Patches and 6Brown dataset outperforms the state-of-the-art descriptors for matching under illumination changes. We also use the new AMOS Patches testing split to evaluate robustness to lighting and season-related conditions. See Table~\ref{tab:results} for results in the matching task, Figure \ref{fig:verification} in the verification task and Figure \ref{fig:amos_test} for comparison on the proposed AMOS Patches test split.

\begin{table}[htb]
    \centering
    \caption{HPatches matching scores (mAP).}
    \begin{tabular}{lccc}
        \toprule
        Training set &\multicolumn{3}{c}{HPatches subset}\\
        & illum & view & full \\
        \midrule
      Liberty & 49.86 & 55.62 & 52.79 \\
      6Brown & 52.39 & 59.15 & 55.83 \\
      PS & 48.55 & \textbf{67.43} & 58.16 \\
      \midrule
      Webcam~\cite{Verdie2015} & 51.82 & 50.77 & 51.29 \\
      AMOS-patches  & 55.17 & 57.94 & 56.58\\
      +Liberty & 56.14 & 60.27& 58.24 \\
      +6Brown & \textbf{56.22} & 61.50& \textbf{58.91}  \\
        \bottomrule
    \end{tabular}
    \label{tab:results}
\end{table}
\begin{table*}[htb]
\centering
\caption{Comparison of the AMOS+6Br HardNet vs. HardNet++~\cite{mishchuk2017working} following the protocol~\cite{wxbs2015}. The number of matched image pairs is shown. The \fbox{numbers} of image pairs in a dataset are boxed. Best results are in \textbf{bold}.}
\setlength{\tabcolsep}{4pt}
\begin{tabular}{lccccccc}
\toprule
&EF~\cite{Zitnick2011}& EVD~\cite{mishkin2015mods}& OxAff~\cite{Mikolajczyk2005} &SymB~\cite{Hauagge2012} 
& GDB~\cite{yang2007registration} & map2photo~\cite{wxbs2015}& LTLL~\cite{Fernando2015}\\


Descriptor & \fbox{33}
 & \fbox{15}
 & \fbox{40}
 & \fbox{46}
 & \fbox{22}
 & \fbox{6}
  & \fbox{172}\\
 
\cmidrule(r){1-8}
HardNet++~\cite{mishchuk2017working} &31 &\textbf{15}  &\textbf{40} &40 &18 &2& \textbf{108} \\
HardNetAMOS+6Br &\textbf{33} &\textbf{15} &\textbf{40}   &\textbf{45} &\textbf{19} &\textbf{4}&106  \\

\bottomrule
\end{tabular}
\label{tab:wxbs-table}
\end{table*}

\paragraph{Wide baseline stereo.} Finally, we evaluate the descriptors on a real-world task -- wide baseline stereo on multiple datasets, following the protocol~\cite{wxbs2015}. Two metrics are reported: the number of successfully matched image pairs and the average number of inliers per matched pair. Results are shown in Table~\ref{tab:wxbs-table}. Edge Foci (EF)~\cite{Zitnick2011}, Extreme view~\cite{mishkin2015mods} and Oxford Affine~\cite{Mikolajczyk2005} benchmarks provide a sanity check --- the performance on the benchmark is saturated and they contain (mostly) images taken from a slightly different viewpoint.

SymB~\cite{Hauagge2012}, GDB~\cite{yang2007registration} and map2photo~\cite{wxbs2015} contain image pairs which are almost perfectly registered, but have severe differences in illumination or modalities, e.g. drawing vs. photo, etc. AMOS+6Br HardNet performs better than baseline HardNet++ on such datasets.
The last dataset -- LTLL~\cite{Fernando2015} consists of historical photos and old postcards. The landmarks are depicted with significant changes in both viewpoint and illumination. Baseline HardNet++ slightly outperforms our descriptor. 
Overall, the benchmark confirms that HardNet trained on AMOS Patches is robust to illumination and appearance changes in real-world scenarios.

\section{Conclusion} 
\label{sec:conclusion}
We provide the AMOS Patches dataset for robustification of local feature descriptors to illumination and appearance changes. It is based on registered images from selected cameras from the AMOS dataset. It has both the training and testing split.

We introduce the local feature descriptor trained on AMOS Patches and 6Brown datasets, which achieves the score of 58.91 mAP in HPatches matching task in full split, compared to the current state-of-the-art: 59.1 mAP (GeoDesc). The advantage of the descriptor is the robustness to illumation. It achieves the state-of-the-art score of 56.22 mAP in matching task, illum split, compared to 52.39 mAP of HardNet++.

We conclude with a list of observations and recommendations related to using webcams for descriptor learning:

\begin{itemize}
    \item Scene parsing methods  do not work well in outdoor webcams. The precision of the near state-of-the-art network \cite{zhou2018semantic} is not satisfactory.
    \item For camera selection we recommend to adopt strict "quality" criteria and be prepared to loose many suitable cameras in the process.
    \item When picking cameras for training manually, a small and diverse subset is better than a bigger one with similar views or imprecise alignment of images.
\end{itemize}

\section*{Acknowledgements}
The authors were supported by the Austrian Ministry for Transport, Innovation and Technology, the Federal Ministry of Science, Research and Economy, and the Province of Upper Austria in the frame of the COMET center SCCH, the CTU student grant SGS17/185/OHK3/3T/13, 
and the OP VVV funded project CZ.02.1.01/0.0/0.0/16\_019/0000765 Research
Center for Informatics.
{\small
\bibliographystyle{ieee}
\bibliography{cvww_template}
}

\end{document}